\documentclass{article}

\usepackage{microtype}
\usepackage{graphicx}
\usepackage{subfigure}
\usepackage{booktabs}

\usepackage{hyperref}
\clearpage{}\usepackage[english]{babel}
\usepackage[utf8]{inputenx}
\usepackage{xspace}
\usepackage{enumerate}

\usepackage[inline]{enumitem}
\usepackage{balance}
\usepackage{textcomp}
\usepackage[shortcuts,acronym]{glossaries}
\usepackage{soul, footnote, xargs}
\usepackage{microtype}

\usepackage{multirow, multicol, booktabs, tabulary, tabu, longtable, array, varwidth}
\usepackage{placeins}
\usepackage[flushleft]{threeparttable}

\usepackage{graphicx, tikz, epstopdf, stfloats, bbding, capt-of}
\usepackage[labelfont=bf]{caption}

\usepackage{amsmath, amsfonts, amssymb, nicefrac}

\usepackage[]{cleveref} \crefname{appsec}{Appendix}{Appendices}
\crefformat{section}{\S#2#1#3}
\crefrangeformat{section}{\S#3#1#4--\S#5#2#6}
\crefformat{subsection}{\S#2#1#3}
\crefformat{subsubsection}{\S#2#1#3}
\crefformat{equation}{(#2#1#3)}
\crefrangeformat{equation}{(#3#1#4--#5#2#6)}
\crefmultiformat{equation}{(#2#1#3)}{ and~(#2#1#3)}{, (#2#1#3)}{ and~(#2#1#3)}
\crefformat{figure}{Fig.~#2#1#3}
\crefrangeformat{figure}{Figs. #3#1#4--#5#2#6}
\crefmultiformat{figure}{Figs.~#2#1#3}{ and~#2#1#3}{, #2#1#3}{ and~#2#1#3}
\crefformat{algorithm}{Alg.~#2#1#3}
\crefformat{table}{Table~#2#1#3}
\crefrangeformat{table}{Tables~#3#1#4--#5#2#6}
\crefmultiformat{table}{Tables~#2#1#3}{ and~#2#1#3}{, #2#1#3}{ and~#2#1#3}

\clearpage{}

\usepackage[preprint]{neurips_2021_ml4ad}

\title{Watch out for the risky actors: {A}ssessing risk in dynamic environments for safe driving}
\clearpage{}

\clearpage{}
\author{
Saurabh Jha, Yan Miao, 
Zbigniew Kalbarczyk, 
Ravishankar K. Iyer\\
University of Illinois at Urbana-Champaign}
\begin{document}

\maketitle
\begin{abstract}
Driving in a dynamic environment that consists of other actors is inherently a risky task as each actor influences the driving decision and may significantly limit the number of choices in terms of navigation and safety plan. 
The risk encountered by the Ego actor depends on the driving scenario and the uncertainty associated with predicting the future trajectories of the other actors in the driving scenario. 
However, not all objects pose a similar risk. Depending on the object's type, trajectory, position, and the associated uncertainty with these quantities; some objects pose a much higher risk than others. 
The higher the risk associated with an actor, the more attention must be directed towards that actor in terms of resources and safety planning. 
In this paper, we propose a novel risk metric to calculate the importance of each actor in the world and demonstrate its usefulness through a case study.  
\end{abstract} \section{Introduction}
\label{s:introduction}
Driving in a real-world environment, with ever-changing dynamics,  consisting of other actors (i.e., non-player character or NPC) is inherently a risky task. 
Each actor in the environment can significantly influence the driving decisions of the Ego actor\footnote{Actor that is being tested and is under the control of the tester.} and limit the number of choices available in terms of navigation and safety plan. 
In extreme cases, these actors, willingly or unwillingly, can thwart the Ego actor's ability to safely complete the task.
Thus, each actor in the environment poses a risk to the Ego actor. 
In this paper, we characterize the risk posed by an actor by estimating it's influence on the Ego actor's decision process in a given environment.
Clearly, the risk is then function of:
\begin{enumerate*}[label=(\roman*)]
    \item the environment (i.e., driving scenario), and 
    \item uncertainty associated with determining the future trajectories by the Ego actor.
\end{enumerate*}

\begin{figure}
    \centering
    \includegraphics[width=0.7\textwidth]{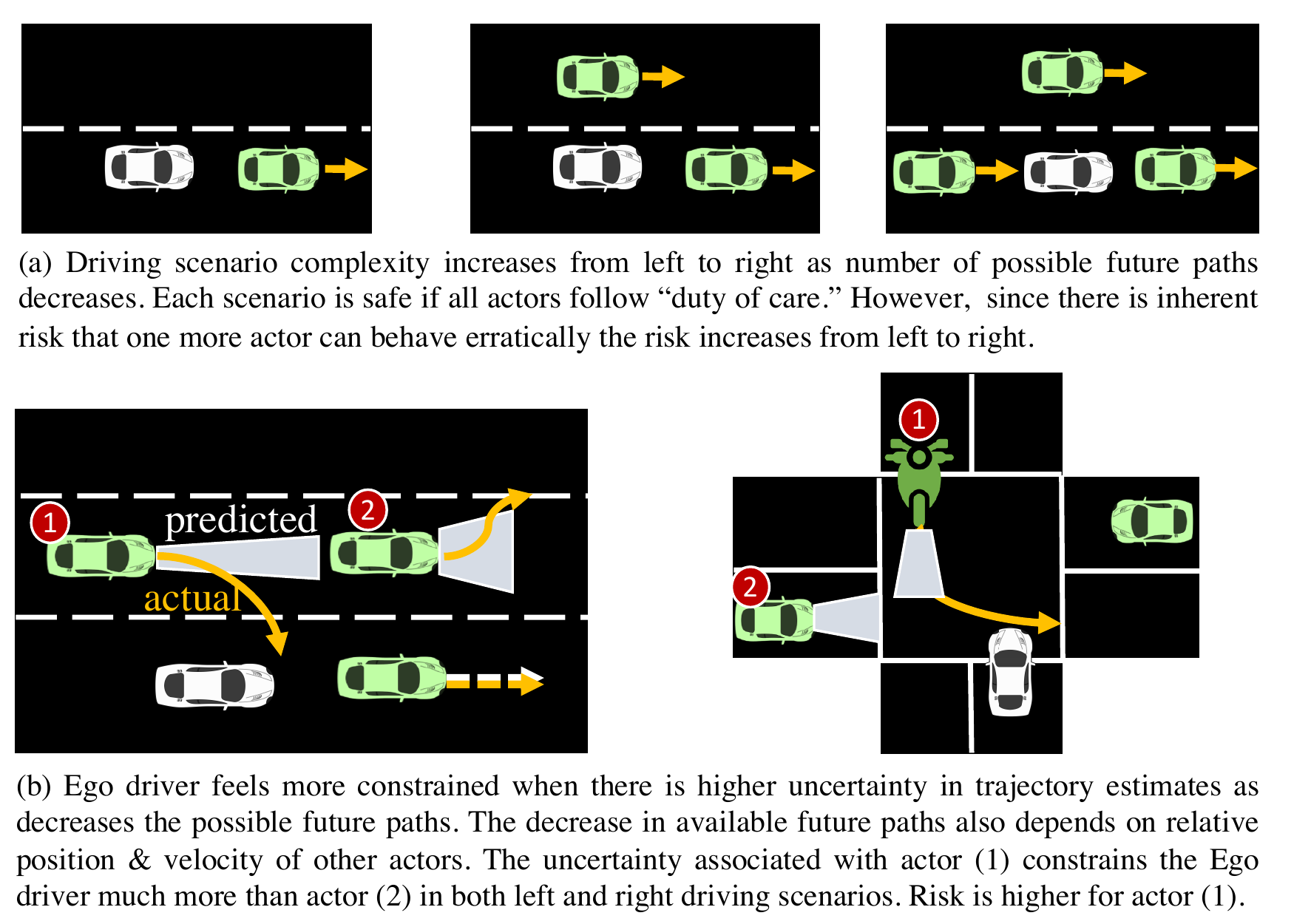}
    \caption{Demonstrating risk with driving scenarios and actors.}
    \label{fig:motivation}
\end{figure}

\textbf{Driving scenario.} Driving scenario is the specification of the initial location of the Ego actor and  trajectories of NPCs. 
With each additional NPC on the road, safe trajectories that can be followed by the Ego actor (i.e., the navigable space) can be significantly restricted.
Similarly, certain locations in the environment are riskier than other as the navigable space may vary significantly from one location to another (e.g., driving in construction zones). 
This decrease in safe trajectories is analogous to the risk because it requires more attentive driving and reduces the fall back options available to the Ego agent in case of unforeseen situations. This is illustrated in \cref{fig:motivation}(a).

\textbf{Uncertainty.}
Risk increases if the future trajectory of the NPCs cannot be predicted accurately by the Ego actor. 
The accuracy of prediction decreases due to unclear intention or sudden change in intention of other participants in the driving environment (e.g., pedestrian decides to suddenly stop or run, emergency braking or breakdown of other vehicles), measurement noise (e.g., due to a bad weather), or imperfections in ML/AI models (e.g., pedestrian is not detected). 
This is illustrated in \cref{fig:motivation}(b)
 
Note that human drivers implicitly and continuously assess the risk associated with both the driving scenario and the uncertainties in the other actor's actions to drive safely. 
Accidents do occur when human drivers fail to assess the risk which can happen when (i) driving under influence or otherwise distracted, (ii) less than fully skilled drivers, and (iii) expectations mismatch because an actor violated the traffic rule or made sudden actions (e.g., braking). 

It is not surprising that the risk is amplified in the case of an AI/ML-operated actors due to aleatoric and epistemic noise.
Hence, it is critically important to develop methods and models to quantify and characterize this risk at runtime to not only ensure safety (by minimizing risk) but also understand/identify situations in which risk is irreducible so as to prepare the passenger for imminent collision via emergency braking or deploying air bags.
In this paper, we demonstrate the practical importance of risk metric and its importance for ensuring safety. The key contributions of this work is in defining a novel risk metric  and demonstrating the use of the proposed  metric to quantify the relative risk associated with each actor on a real-world driving scenario. 
    
Although, in this paper, we focus only on self-driving cars as a use case, our framework 
is general and can be applied to other autonomous systems such as aerial vehicles or ML-driven computer systems.

\textbf{Related work.} There are mainly two lines of research associated with defensive driving: (i) identifying safe distance from other actors assuming everyone follows the  ``duty-of-care'' policies, (ii) identifying out-of-training-distribution (OOD) scenarios so as to plan for the worst-case for avoiding collisions. 
\textit{Duty-of-care approaches} such as Safety Force Field (SFF)~\cite{SafetyForceField} and Responsibility-Sensitive Safety (RSS)~\cite{shalev2017formal} are geared towards estimating the safe distance from other actors assuming that everyone follows rules of the road. Additionally, the goal is to identify the culprit in case of a safety hazard. 
\textit{OOD} models such as \cite{pmlr-v119-filos20a} are geared towards identifying driving scenarios in which the Ego actor's future trajectory (i.e., plan) has significant variance. Such variance can be characterized by using ensemble of models or diverse data.  
Under high variance, the Ego actor chooses to use the most pessimistic plan in order to avoid collisions.
For example, in \cite{pmlr-v119-filos20a}, Ego actor's plan is generated using Bayesian imitative model, and the variance of the imitation prior with respect to the model posterior is used as a proxy for identifying distribution shifts.
On detecting distribution shift, the Ego actor can either plan for the worst-case model or the average model. 
Finally, in another line of research~\cite{Philion_2020_CVPR}, authors  quantify the impact of uncertainty in driving performance, but do not quantify the quality of the generated plan in terms of its safety.
In contrast, our goal is to (i) assign risk associated with an actor or a driving scenario by characterizing it's negative influence on the degree of freedom of the Ego actor's navigational choices, (ii) quantify the risk associated  with the current plan and proactively mitigate safety hazards. 
Thereby, allowing us to identify interesting driving scenarios in offline mode, and to select the most safe plan help the Ego actor drive defensively.
 \section{Problem Setting and Notation}
\label{s:problem-def}
The current approaches in identifying safety-critical actors on the road relies {\it only} on forward simulating techniques to identify collision sets/trajectories. 
However, these techniques do not account for the attention required for driving in strenuous driving scenarios where the Ego actor has limited number of choices/plans. 

The goal our work is to quantify the risk encountered by the Ego actor with respect to an actor on the road for a given driving scenario. 
Towards that end, we the propose a risk metric which resembles human-driven intuition and reasoning of safe driving where the goal is to navigate such that the actors risk is reduced.
Our metric is inspired from ablation studies and more specifically from Barlow's importance metric~\cite{barlow1975importance}. 
We demonstrate the use of the proposed metric on a Physics-based simulator of self -driving cars. In our experiments, we assume that the Ego actor uses ensembles of modules (and associated models) responsible for perception (the object detection, tracking, and sensor fusion), trajectory planning and control.

In this work, we ignore issues of: (i) \textit{fragmentation} caused by model's inability to match the detected actor to its trajectory, and (ii) \textit{false detection} that leads to appearance of ghost objects in the world as perceived by the Ego actor.

These are benign assumptions for many applications in
robotics. If required, these quantities can also be learned
from data.

\subsection{Determining risk posed by an actor}
We first describe the risk in terms of importance of an actor. In this setting, we assume that the the state of each actor is known ahead of time and has no uncertainty. Finally, we discuss the risk calculation for driving scenarios in which the state of the actors are uncertain. 

\subsection{Determining risk with ground truth data}
Let us assume that there are $N$ actors in the world including the Ego actor.
Let us denote the state of an actor $i$ at time $t$ by $x^{(i)}_t \in \mathbb{R}^{3}$, and the trajectory (i.e., trace of the actor's state) from time $t$ to $t+k$ given by $X^{(i)}_{t,k}$.
Let us denote the trajectory of all actors except the Ego actor from time $t$ to $t+k$ by $\mathbb{X}_{t,k} = {X^{(1)}_{t,k}, ..., X^{(N)}_{t,k}}$.
We can now describe the driving scenario ($\mathbb{S}$) of length $T$ as a tuple consisting of a map ($\mathbb{M}$), trajectories of all the actors except the Ego actor, and the initial position of the Ego actor ($x^{ego}$):
\begin{equation}
    \mathbb{S} = <\mathbb{M}, \mathbb{X}_{0,T}, x^{ego}_{t=0}>
\end{equation}

In this paper, we assume that we have access to a planner and router ($f_p$) which uses the trajectories of all the other actors from time $t$ to $t+k$ ($\mathbb{X}_{t, k}$) and the position of the Ego actor at time $t$ ($x^{ego}_t$) to generate a set of future trajectories ($\mathbb{Z}^{ego}_{t,k}$) that the Ego actor can follow safely from time $t$ to $t+k$ while following all the rules of the road.

\begin{equation}
    Z_{t,k} = f_p(\mathbb{M}, \mathbb{X}_{t, k}, x^{ego}_{t})
    \label{eq:planner}
\end{equation}
If we assume that there are no actors in the world/driving scenario, we get a set consisting of all the navigable future trajectories.
\begin{equation}
    Z^{\emptyset}_{t,k} = f_p(\mathbb{M}, x^{ev}_{t=0})
    \label{eq:navigable_space}
\end{equation}
We now define total risk ($\rho_{t,k}$) encountered by an Ego actor in a driving scenario as the normalized reduction in future trajectories due to presence of all actors on the road. 

\begin{equation}
    \rho_{t,k} = \frac{Z^{\emptyset}_{t, k} - Z_{t, k}}{Z^{\emptyset}_{t, k}} 
    \label{eq:totalrisk}
\end{equation}

Similarly, risk encountered by the Ego actor due to a specific actor $i$ is given by:
\begin{equation}
    \rho^{i}_{t,k} = \frac{Z_{t, k} - Z^{/i}_{t, k}}{Z^{\emptyset}_{t, k}} 
    \label{eq:actorrisk}
\end{equation}

However, estimating \cref{eq:totalrisk} and \cref{eq:actorrisk} is difficult at runtime as motion-planning is PSPACE-hard, which is an indication of the computational intractability in the degrees of freedom of the agent~\cite{SCHWARTZ1983298, canny1988complexity}. 
Therefore, existing planner such as RRT* uses sampling techniques to generate the most optimal plan with a fixed time budget~\cite{lavalle1998rapidly}. 
Therefore, we approximate $\rho_{t,k}$ posed by an actor $i$ in a driving scenario as the change in the future trajectories considering all actors with and without actor $i$. 
Let us denote this alternate definition by ($\gamma^i$) given by \cref{eq:risk_alternative}.

\begin{align}
    \gamma^{(i)}_{t,k} &= |f_p(X^{(1)}_{t,k},...., X^{(N)}_{t,k}) - f_p(X^{(1)}_{t,k},..., X^{(i-1)}_{t,k}, X^{(i+1)}_{t,k}, X^{(N)}_{t,k} )|\\
    &= |Z_{t,k} - Z^{/i}_{t,k}|
    \label{eq:risk_alternative}
\end{align}
Depending on the implementation of the planner, the difference operator in \cref{eq:risk_alternative} can be KL-divergence (in case the planner outputs a distribution) or euclidean difference (in case the planner outputs one single trajectory) between the two generated paths normalized by the number of waypoints in the trajectory.

\subsection{Determining risk at runtime}
At runtime, the Ego actor does not have access to the ground truth data. Hence, the Ego actor must estimate the risk with noisy measurement and prediction of future trajectories of other actors. Without loss of generality, it can be shown that uncertainty can be incorporated in  \cref{eq:risk_alternative}. For example, we can calculate the expected risk and its variance on the sampled trajectories from a distribution,  $\bar{X}^{(i)}_{t, t+k}$, which models the uncertainty in the estimates of future trajectories of all actors.

 \section{Case Study}

\begin{figure}
    \centering
    \includegraphics[width=\textwidth]{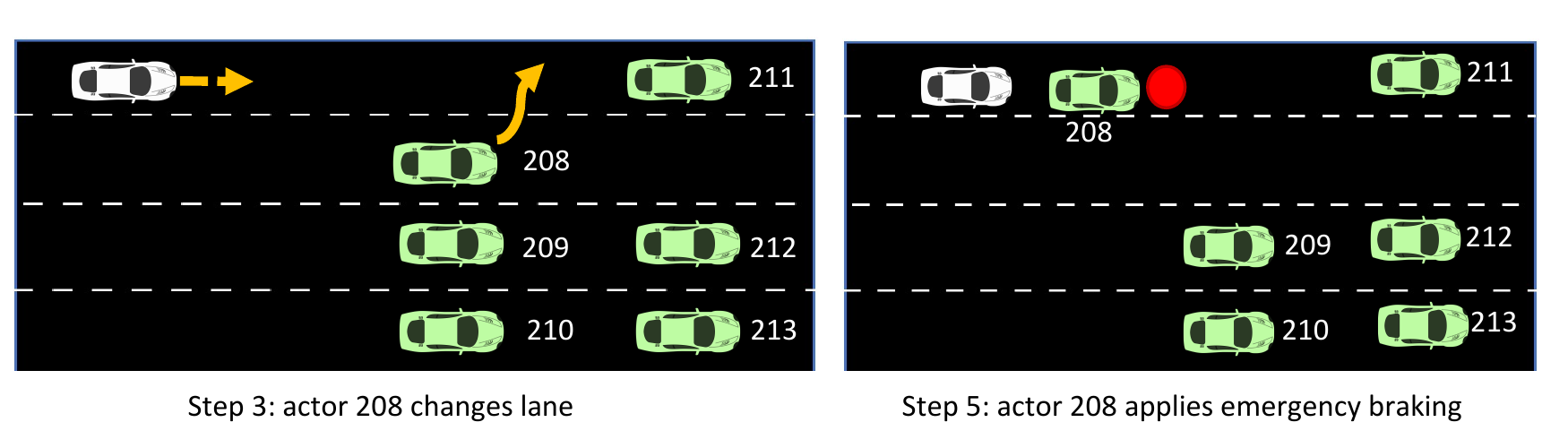}
    \caption{Case study driving scenario}
    \label{fig:case_study}
\end{figure}
We created an example driving scenario consisting of seven actors, including the Ego actor as shown in \cref{fig:case_study}.
The Ego actor controls the white car, and the simulator controls the green cars. 
The scenario consists of five steps ---  \textit{s1: initialization, s2: steady driving, s3: lane change, s4: steady driving, and s5: emergency braking}. 
In the \textit{s1: initialization} step, all actors are initialized. These actors accelerate to reach a constant velocity. 
In the \textit{s2: steady driving} step, all actors maintain a constant velocity.
In the \textit{s3: lane change} step, actor 208 changes the lane as shown in the figure, while others continue to maintain the constant velocity.
In the \textit{s4: steady driving} step, actor 208 has completed the lane change. All actors maintain a constant velocity.
In the \textit{s5: emergency braking} step, actor 208 applied emergency braking.
s1 is run from timestep 0--375, s2 is run from timestep 375--795, s3 is run from timestep 795--1035, 
s4 is run from timestep 1035--1215, 
and s5 is run from 1215--1905.
The variable timing is because scenario generation is event-triggered  instead of time-triggered. 

In our experiment, we use PyLOT\footnote{\url{https://github.com/erdos-project/pylot}} to control the Ego actor and Carla~\cite{carla17} for the simulation. 
PyLOT is configured to use perfect detection, perfect tracking, linear behavior prediction model, and RRT* planner with default parameters. 
In our case study, we observe that the Ego agent (which uses RRT* planner) decides to follow object 211 until s4, and at s5, it decides to change the lane and follow actor 209.

Currently, PyLOT lacks the capability for estimating uncertainty associated with future trajectories. Therefore, we only consider a single trajectory estimate of each actor as provided by PyLOT. 
Despite these limitations, we discuss the validity of our proposed metric and its use in driving safely. 

\subsection{Results}

In this experiment, we use a perfect tracker; i.e., we directly use the ground truth data on tracking from the simulator, thus eliminating the need for object detection. However, the behavior prediction model used is linear. 

\begin{figure}
    \centering
    \includegraphics[width=\textwidth]{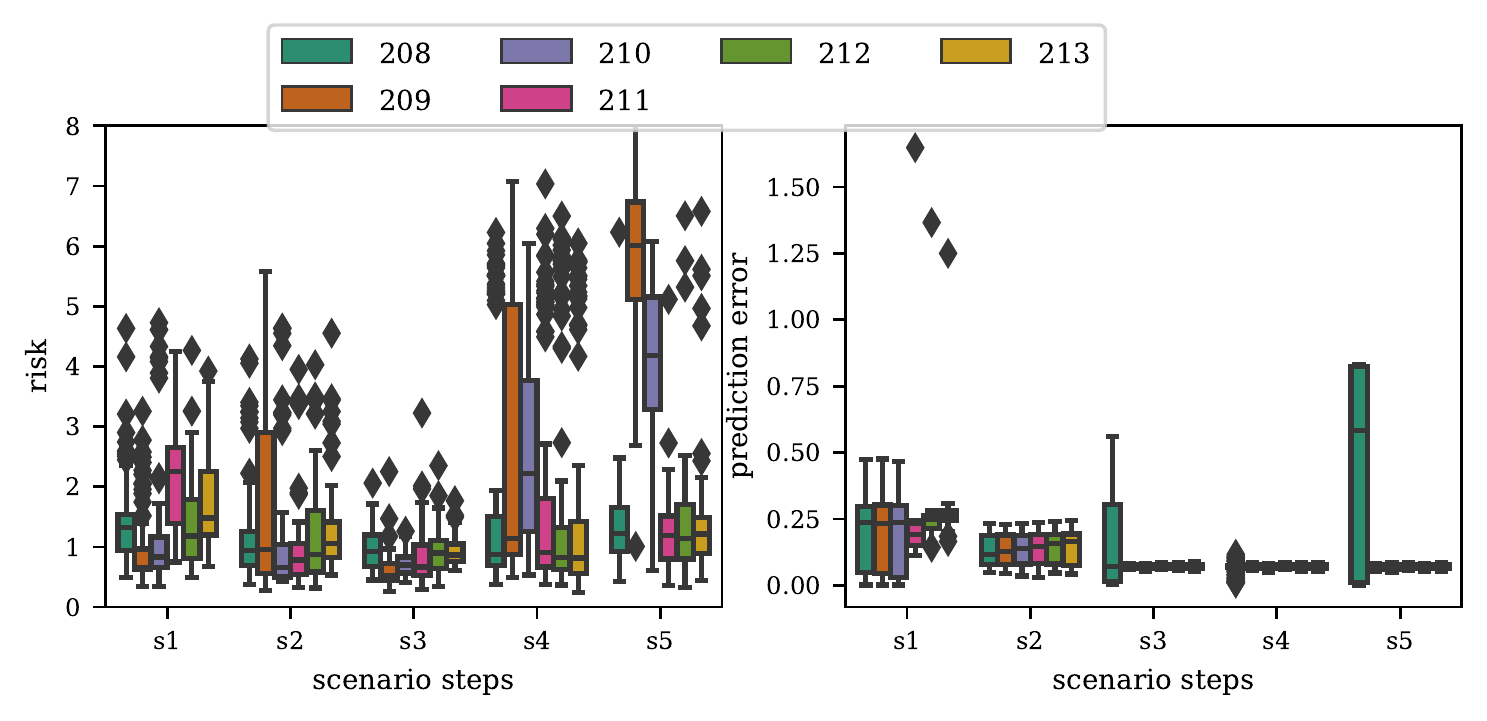}
    \caption{Risk score and prediction error associated with each actor at each driving scenario step when using \textbf{perfect tracking}.}
    \label{fig:hist}
\end{figure}

Initially, all actors are accelerating to attain their target speed; hence, each actor has a relatively higher prediction error. However, as time passes, the prediction error decreases for each actor except actor 208.
The prediction error in steps s3 and s5 for actor 208 increases significantly (see right subfigure in \cref{fig:hist}).
This is due to the fact that actor 208 makes a sudden change in each of the step, which cannot be determined using the linear model-based behavior prediction.
These sudden changes in the environment are generally hard to predict irrespective of the underlying behavior prediction model; thus, leading to uncertainty in the future trajectory estimates. 

As discussed before, PyLOT lacks the capability for estimating uncertainty associated with future trajectories. Hence, we evaluate the risk by calculating the median of the importance score (which is calculated using \cref{eq:risk_alternative}) calculated for each driving scenario step for each actor, where we use the PyLOT estimated trajectories instead of the ground truth data. 
The risk for each actor at each step is shown in \cref{fig:hist} (left subfigure).
We observe that the risk depends on the Ego actor's planned trajectory. 
In s1, the Ego actor frequently changes its planned future trajectory, and hence, objects that are in its path become important depending on the chosen trajectory. 
At s2, the median risk associated with each decreases to less than one. However, there is higher variability in risk for actor 209 as in some planning iterations the Ego actor decides to change lane go behind actor 209. 

At s3, since the environment is stable, the Ego actor decides to remain behind actor 208, and hence the median risk for actor 208 is higher than others. 
At s4, the Ego actor now switches its decision to go behind either actor 209 or 210; giving more preference to actor 210. The median risk for the actor 210 and 209 increases to 2.2 and 1.1 respectively. The variability is also higher than other as seen from the box plot.  Other actors remains to be less riskier than 209 and 210.
At s5, even though the prediction error for actor 208 increases (median of 0.50; see box plot), the risk increases marginally from s4 to s5.  The risk associated with actor 208 is much lower compared to the risk associated with actors 209 (risk score 6) and 210 (risk score 4). 
Clearly, the risk associated with each actor changes with time and at any given time only few actors are more important than others. 
In our scenario, although the Ego actor is behind actor 208 for the most part, the risk is relatively lower than actor 209 (whose risk score is highest at S5). This is because the Ego actor has decided to switch lanes before the emergency braking of actor 208. Thus, risk also depends on the Ego plan because of our chosen approximation (refer to \cref{eq:risk_alternative})

\begin{figure}
    \centering
    \includegraphics{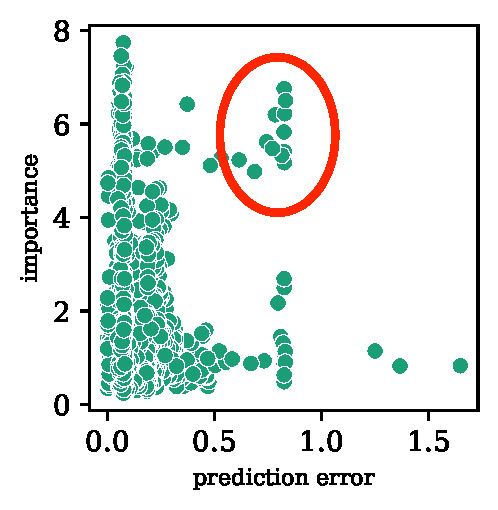}
    \caption{Relationship between importance of an actor in path planning i.e.,\cref{eq:risk_alternative} and the prediction error.}
    \label{fig:perfect_scatter}
\end{figure}

This leads us to conclude that risk posed by an actor is a function of its state (i.e., future trajectory), the prediction error\footnote{At runtime, one cannot determine prediction error but it can be approximated by estimating the uncertainty associated with the predicted future trajectories of the actors.  
} and the Ego plan. 
\cref{fig:perfect_scatter} captures this intuition.
Each point on this scatter plot represents an actor with a corresponding importance score \cref{eq:risk_alternative} and prediction error at timestep $t$ in the driving scenario.
Points inside the circle are riskier than others. Proactive mitigation techniques such as planning a path that reduces the risk associated with the actors in the highlighted region will significantly improve the safety. In our case study, we find that such risk-aware proactive mitigation avoids collision.

 \section{Future Work}
\label{s:future}

In future, we plan to use this metric in the following ways:
\begin{enumerate}[label=(\alph*)]
    \item \textit{Curating standard driving scenario benchmarks}: Since the metric explicitly focuses on identifying strenuous driving scenarios and actors, it is possible to use the proposed metric to create a standard benchmark by mining difficult to navigate driving scenarios using publicly available datasets such as BDD~\cite{yu2018bdd100k} and nuScenes dataset~\cite{nuscenes2019}.
    
    \item \text{Safety hazard mitigation}: We plan to use the proposed metric to identify hazardous driving situations so as to disengage or apply emergency brakes. 
    
    \item \textit{Risk-aware compute resource utilization}: Since at any given point in time, only some actors pose a high risk, we plan to use our metric for deciding on how to allocate compute resources so as to generate robust predictions and plans. 
    
    \item \textit{Risk-aware planning}: We plan to develop methods that will enable us to predict the future risk associated with each actor to navigate safely. The objective of such a planner will be to minimize risk while maximizing reward (i.e., reaching the destination).  
\end{enumerate}

We also plan to demonstrate our metric on more realistic agents whose planners not only consider obstacles but rules of the road and maps for navigation. 

\section{Conclusion }
\label{s:conclusion}

Driving in a dynamic environment that consists of other actors is inherently a risky task as each actor influences the driving decision and may significantly limit the number of choices in terms of navigation and safety plan. 
However, not all objects pose a similar risk. Depending on the object's type, trajectory, position, and the associated uncertainty with these quantities; some objects pose a much higher risk than others.
In this paper, we proposed a novel risk metric to calculate the importance of each actor in the world under uncertainty and demonstrate its usefulness through a case study. 

 \newpage
\bibliography{main.bib}
\bibliographystyle{icml2021}

\end{document}